\documentclass[letterpaper, 10 pt, conference]{ieeeconf}  

\IEEEoverridecommandlockouts                              

\usepackage{times}
\usepackage{soul}
\usepackage{url}
\usepackage[hidelinks]{hyperref}
\usepackage[utf8]{inputenc}
\usepackage[small]{caption}
\usepackage{graphicx}
\usepackage{amsmath}
\usepackage{amssymb}
\usepackage{amsfonts}
\usepackage{booktabs}
\usepackage{algorithm}
\usepackage[switch]{lineno}
\usepackage{cite}
\usepackage{textcomp}
\usepackage{xcolor}
\usepackage[nolist]{acronym}
\usepackage{placeins}
\usepackage{bm}
\usepackage{enumerate}
\usepackage{balance}
\usepackage{makecell}
\usepackage{comment}
\usepackage{algpseudocode}
\usepackage{multirow}
\usepackage{float}

\title{\LARGE \bf
Feature Aggregation with Latent Generative Replay for Federated Continual Learning of Socially Appropriate Robot Behaviours
}

\ifx\blindsubmission\undefined

    \author{Nikhil~Churamani$^{1,*}$, Saksham~Checker$^{2}$, Fethiye~Irmak~Dogan$^{1}$, Hao-Tien~Lewis~Chiang$^{3}$,  Hatice~Gunes$^{1}$%
    \thanks{$^{1}$N.~Churamani, F.~I.~Dogan, and H.~Gunes are with the Department of Computer Science and Technology, University of Cambridge, UK.}
    \thanks{$^{2}$S.~Checker is with King's College London, UK. S.~Checker contributed to this work while undertaking a remote visiting studentship at Department of Computer Science and Technology, University of Cambridge.}
    \thanks{$^{3}$H.T.L.~Chiang is with Google DeepMind.}
    \thanks{This work is supported by Google under the GIG funding scheme. The authors also thank Jie~Tan and Carolina~Parada for their valuable feedback.}
    \thanks{$^{*}$Corresponding Author: {\tt Nikhil.Churamani@cl.cam.ac.uk}}
    \thanks{Code: \url{https://github.com/nchuramani/FedLGR_SocRob}}
}    
\else
  \author{{Anonymous Authors}
    \thanks{Affiliations Removed for double-blind submission.}
    \thanks{Acknowledgement removed for double-blind submission.}
    \thanks{Code: Link removed for double-blind submission.}
    }

\fi

\begin{document}
\setlength{\belowcaptionskip}{-12pt}

\maketitle

\begin{abstract}

It is critical for robots to explore \acf{FL} settings where several robots, deployed in parallel, can learn independently while also sharing their learning with each other.
This collaborative learning in real-world environments requires social robots to adapt dynamically to changing and unpredictable situations and varying task settings. 
Our work contributes to addressing these challenges by exploring a simulated living room environment where robots need to learn the social appropriateness of their actions. First, we propose \acf{FedRoot} averaging, a novel weight aggregation strategy which disentangles feature learning across clients from individual task-based learning. 
Second, to adapt to challenging environments, we extend  \ac{FedRoot} to \acf{FedLGR}, 
a novel \acf{FCL} strategy that uses \ac{FedRoot}-based weight aggregation and embeds each client with a \textit{generator} model for pseudo-rehearsal of learnt feature embeddings to mitigate forgetting in a resource-efficient manner. 
Our results show that \ac{FedRoot}-based methods offer competitive performance while also resulting in a sizeable reduction in resource consumption (up to 86\% for CPU usage and up to 72\% for GPU usage).
Additionally, our results demonstrate that \ac{FedRoot}-based \ac{FCL} methods outperform other methods while also offering an efficient solution (up to 84\% CPU and 92\% GPU usage reduction), with \ac{FedLGR} providing the best results across evaluations.

\end{abstract}

\begin{figure*}[t!]
    \centering
    \includegraphics[width=\textwidth]{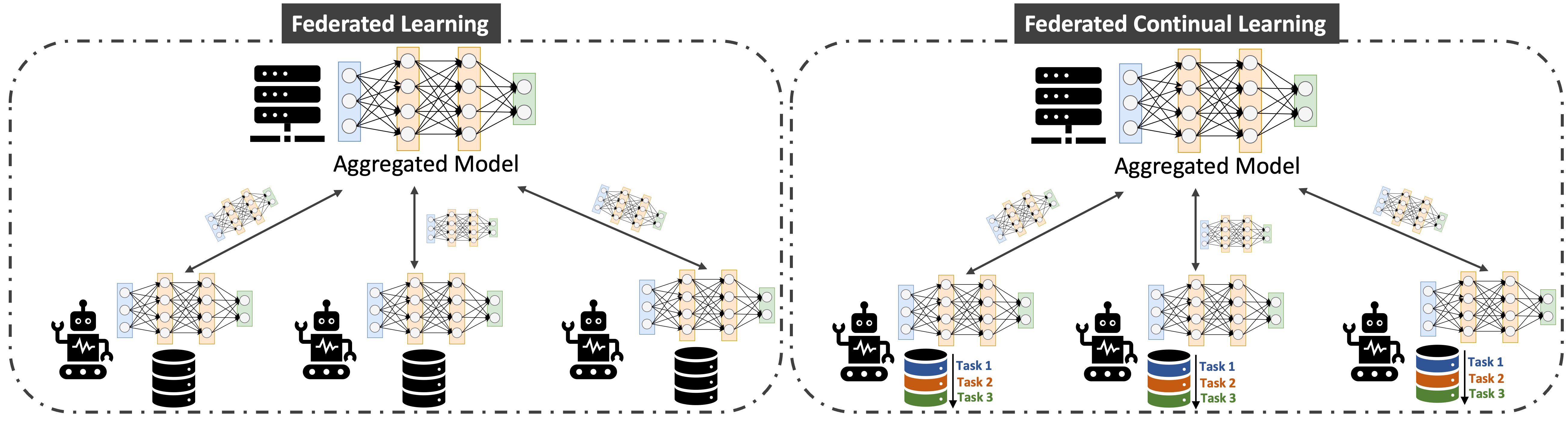}
    \caption{\textbf{\acf{FL}} (left): Local models are aggregated on the server without sharing data. \textbf{\acf{FCL}} (right): Individual robots incrementally learn tasks, periodically sharing model updates with each other.}
    \label{fig:learning}
\end{figure*}

\section{Introduction}

As advances in \acf{AI} gear robots towards a ubiquitous application, they are expected to be deployed across environmental and contextual settings, interacting with several users at a time and learning different tasks~\cite{Guerdan2023FCL}. Each robot, operating in its unique application, should be able to adapt to the dynamics of its ever-changing environment~\cite{Churamani2020CL4AR} while also being sensitive to the changing preferences of the users it interacts with~\cite{dogan2024grace}. Such an understanding of social dynamics and norms~\cite{Ayub2020What} can help robots effectively navigate complex social settings while also offering enriching interaction experiences to their users. Given the vast array of such potential applications~\cite{MAHDI2022104193}, robots can benefit from sharing knowledge and individual experiences with one another. For complex and diverse applications, there is a need to move beyond centralised platforms towards more \textit{distributed learning} paradigms, enabling robots to keep learning \textit{continually} while sharing their learning with others. 

Moving beyond \textit{centralised} learning paradigms, where individual robots only gather data and send it to a central server to be aggregated and used to train a single global model for application, \acf{FL}~\cite{McMahan2017FL} (see Fig.~\ref{fig:learning};~left) offers an efficient solution where individual robots can learn \textit{independently} from their own \textit{unique experiences}, updating their learning models using only the data collected by them \textit{locally}. Over time, these local updates can be \textit{aggregated} across the centralised server in the form of model updates that can inform the training of the unified global model. 
Recently, \ac{FL} has been explored for robotic and autonomous systems~\cite{Xianjia2021FL, Guerdan2023FCL}, allowing for collaborative learning across robots, learning from each others' experience while maintaining end-user privacy. Even under \ac{FL} settings, there might be `information/data leak'~\cite{Carlini2023Extracting} over time as learnt features and task-based learning layers are shared with the server, which may reveal how end-users interact with individual robots. Further, as the robots may be equipped with several hierarchical learning layers, aggregating individual model weights into a global model can be computationally expensive~\cite{Guerra2023Cost}. In such cases, splitting model learning into parts can allow robots to effectively share their learning without violating end-user privacy. 

A key challenge for robots in \ac{FL} settings within dynamic human environments is to effectively discern novel information from past experiences and adapt their learning models to accommodate new knowledge. 
As the real world changes over time~\cite{Hadsell2020CL}, each individual robot may learn with incremental and/or sequential streams of data where data might not be \acf{i.i.d}. This may result in forgetting past knowledge and overfitting to the current task or data. Although \ac{FL} allows for model aggregations across client robots, such forgetting locally will cause the globally aggregated model to forget past knowledge. \ac{CL}~\cite{Hadsell2020CL} can help address this problem by enabling individual robots to adapt their learning with incrementally acquired data from non-stationary or changing environments~\cite{Churamani2020CL4AR}. This allows robots to accumulate new information locally while preserving previously seen knowledge. Combining the principles of \ac{FL} and \ac{CL}, \acf{FCL}~\cite{Yoon2021FCL, Guerdan2023FCL} (see Fig.~\ref{fig:learning};~right) can allow individual robots to learn over time in their unique settings, and benefit from other robots' learning. 

\begin{figure}[t]
    \centering
    \includegraphics[width=\columnwidth]{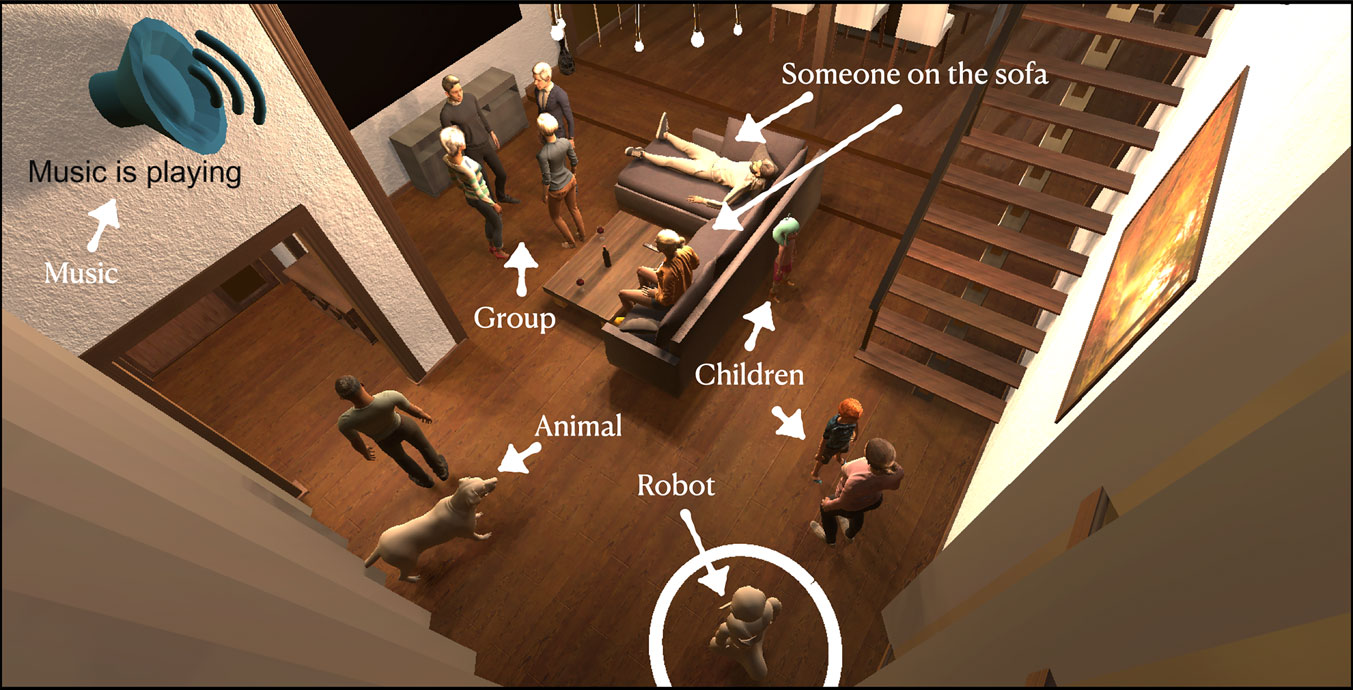}
    \caption{MANNERS-DB:~A Living Room scenario with a robot~\cite{tjomsland2022mind}.}
    \label{fig:mannerdb}
\end{figure}

Addressing both \ac{FL} and \ac{FCL} objectives as described above, in this paper, we propose an end-to-end novel approach using simulated home environments where the robot is required to learn the social appropriateness of different tasks from scene renders. To achieve this, we explore the MANNERS-DB dataset~\cite{tjomsland2022mind} which 
provides social-appropriateness ratings for different robot actions in simulated living room settings (see Fig.~\ref{fig:mannerdb}). 
We first propose \acf{FedRoot} averaging, a novel \ac{FL} strategy which disentangles feature-based learning across clients from individual task-based learning. Under \ac{FedRoot}, only the feature extraction layers are aggregated between \textit{clients}; keeping task-based learning \textit{strictly local}, enhancing resource efficiency and privacy for \ac{FL}. Adapting popular \ac{FL} strategies to use \ac{FedRoot} instead, we present a novel \ac{FL} benchmark for learning the social appropriateness of different robot actions in diverse social configurations. 
Next, applying \ac{FedRoot} to \ac{FCL} settings, we propose \acf{FedLGR}, a novel \ac{FCL} strategy that combines \ac{FedRoot} with \acs{LGR}~\cite{Stoychev2023LGR} by adding a local \textit{generator} model to the client for efficient pseudo-rehearsal of learnt features to mitigate forgetting. For successive tasks, pseudo-samples of previously seen tasks are generated and mixed with the current task's data to simulate \textit{i.i.d} settings. We benchmark \ac{FedLGR} against popular \ac{CL} adaptations of \ac{FL} strategies, with and without \ac{FedRoot}-based aggregation, to learn socially appropriatenes of robot behaviours in a task-incremental manner. Our results show that both \ac{FedRoot} and \ac{FedLGR} reduce the memory and resource footprint of their \ac{FL} and \ac{FCL} counterparts without compromising model performance.



\section{Related Work}
\subsection{Socially Appropriate Robot Behaviours}
Social interactions are governed by varying norms, laying out expected behaviours from individuals that may be considered socially appropriate by others~\cite{Roesler2022Socially}. For social robots operating in human-centred environments, where interaction contexts and individual user preferences can vary widely, these robots should be able to comprehend and respond to the unique dynamics of each contextual setting~\cite{Li2019Perceptions}. This not only enhances their ability to navigate complex social situations but also improves users' perception of the robots and their acceptance in social \ac{HRI}~\cite{Roesler2022Toward}. Furthermore, social robots need to \textit{continually} learn from their interactions and adjust their behaviours accordingly~\cite{Shiarlis2017Acquiring}. Whether it is effectively navigating complex social environments~\cite{francis2023principles}, learning approach and positioning behaviours~\cite{gao2019learning, McQuillin2022RoboWaiter} or learning task-specific behaviours~\cite{tjomsland2022mind}, it is essential for robots to consider the social-appropriateness of their behaviours in order to comply with social norms~\cite{Ayub2020What}. With distributed deployment increasingly becoming a reality, there is a need to investigate \textit{federated} application frameworks that allow robots to learn in their own unique environments while also informing a global learning of \textit{generalisable} social norms and preferences~\cite{Guerdan2023FCL}. Our work contributes to this direction in a more resource-efficient and privacy-preserving manner.

\subsection{Federated Learning}
\label{sec:fl}
Distributed learning settings can be particularly desirable for social robotsso that they can understand and learn socially appropriate behaviours that are shaped based on the context of the interaction, environmental factors as well as individual user preferences~\cite{Guerdan2023FCL}. Towards this goal, \acf{FL}~\cite{McMahan2017FL} enables a network of distributed client devices (such as robots) to learn individually with locally gathered data while updating their model parameters by aggregating a global learning model (server) that combines these updates from each individual device, over successive update rounds and distributes this model back to individual clients. 
This way, each client shares their knowledge with the other clients without sharing any local data, and it receives updates from a global model that combines the knowledge from other clients. 
Strategies such as \acf{FedAvg}~\cite{Li2020FedAvg} offer a straightforward approach for weight aggregation by collating individual client model weights and computing a weighted average to form the global model. However, \ac{FedAvg} is sensitive to data imbalances and concept drifts, especially in non-\ac{i.i.d} data settings. Several improvements have been proposed on \ac{FedAvg}, for instance, FedBN~\cite{Li2021FedBN} that adapts \ac{FedAvg} by keeping the parameters for all the \textit{BatchNorm} layers `strictly local', that is, all other model weights are aggregated across clients apart from the \textit{BatchNorm} parameters. Similar to FedBN, FedProx~\cite{Li2020FedProx} also proposes improvements over \ac{FedAvg} by allowing for only partial aggregation of weights by adding and tracking a \textit{proximal term} to \ac{FedAvg}. FedOpt~\cite{Reddi2021FedOpt} offers a `general optimisation framework' where each client uses a \textit{client optimiser} to optimise on local data while the server updates apply a gradient-based \textit{server optimiser} to the aggregated model weights. FedDistill~\cite{Jiang2020FedDistill}, on the other hand, aims to improve the ability of clients to deal with \textit{heterogeneous} data conditions by using \textit{knowledge distillation}~\cite{Hinton2015Distillation}. Each client maintains two models: (i) a local copy of the global model and (ii) a personalised model that acts as a teacher to the student global model. The updated student model is then aggregated across clients. In this work, we adapt these methods for a regression-based task, learning to predict the social appropriateness of different robot actions. 

\subsection{Federated Continual Learning}
\label{sec:fcl}
As robots in real-world settings continually encounter novel information, under non-\ac{i.i.d} conditions, their ability to remember previously learnt tasks may progressively deteriorate, resulting in catastrophic forgetting~\cite{LESORT2020CL4R}. \acf{CL}~\cite{Hadsell2020CL, Parisi2018b} strategies may enable robots to learn and adapt throughout their `lifetime', balancing \textit{incremental} learning of novel information with the retention of past knowledge. 
\acf{FCL}~\cite{Yoon2021FCL} combines \ac{FL} and \ac{CL} principles, enabling individual robots to \textit{incrementally} learn \textit{without forgetting} on streams of gathered data. Individual robots learn a series of local tasks while periodically updating the parameters of a global \textit{aggregated} model that combines updates from individual clients over successive update rounds. Several recent works~\cite{Yoon2021FCL} explore \ac{FCL} for vision~\cite{Shenaj_2023_CVPR} and natural language processing~\cite{chaudhary2022federated} applications, however very little work has been done to explore its application for social robots~\cite{Guerdan2023FCL}. Adapting \ac{FL} approaches by adding \ac{CL}-based objectives can offer straightforward solutions for \ac{FCL}. For instance, regularisation-based methods such as \acf{EWC}~\cite{kirkpatrick2017overcomingEWC}, \ac{EWC}Online~\cite{schwarz2018progressEWCOnline}, \ac{SI}~\cite{zenke2017continualSI} or \ac{MAS}~\cite{aljundi2018memoryMAS} can be used to apply penalties on weight updates between old and new tasks to help mitigate forgetting. 
Rehearsal strategies such as \ac{NR}~\cite{Hsu18_EvalCLNR} can be used to maintain local memory buffers for each robot to store and rehearse previously seen data to preserve knowledge. Alternatively, an efficient pseudo-rehearsal of data~\cite{Shin2017DGR} or features~\cite{Stoychev2023LGR, Liu_2020_CVPR_Workshops} can help mitigate forgetting by maintaining generators that model local data distributions for previously seen tasks. In this work, we augment \ac{FL} with regularisation-based \ac{CL} objectives for multi-label regression, predicting the social appropriateness of the different robot actions under different contextual settings.

\section{Methodology}

This section presents our end-to-end approach that enables robots to learn the social appropriateness of different behaviours under in a continual manner under federated settings.
First, \ac{FedRoot}, a novel \ac{FL} strategy is proposed to keep task-based learning \textit{strictly local} by only sharing the feature extraction layers between clients. This avoids revealing individual preferences in terms of what may be considered appropriate for the robot to do under different social configurations, for instance, involving children, elderlies or animals, ensuring end-user privacy. Next, \ac{FedRoot} is extended to \ac{FedLGR} to enable continual learning over incrementally acquired task data employing a generative feature rehearsal approach for a resource-efficient pseudo-rehearsal~\cite{vandeVen2020,Stoychev2023LGR}. 

\begin{figure}[t!]
    \centering
    \includegraphics[width=\columnwidth]{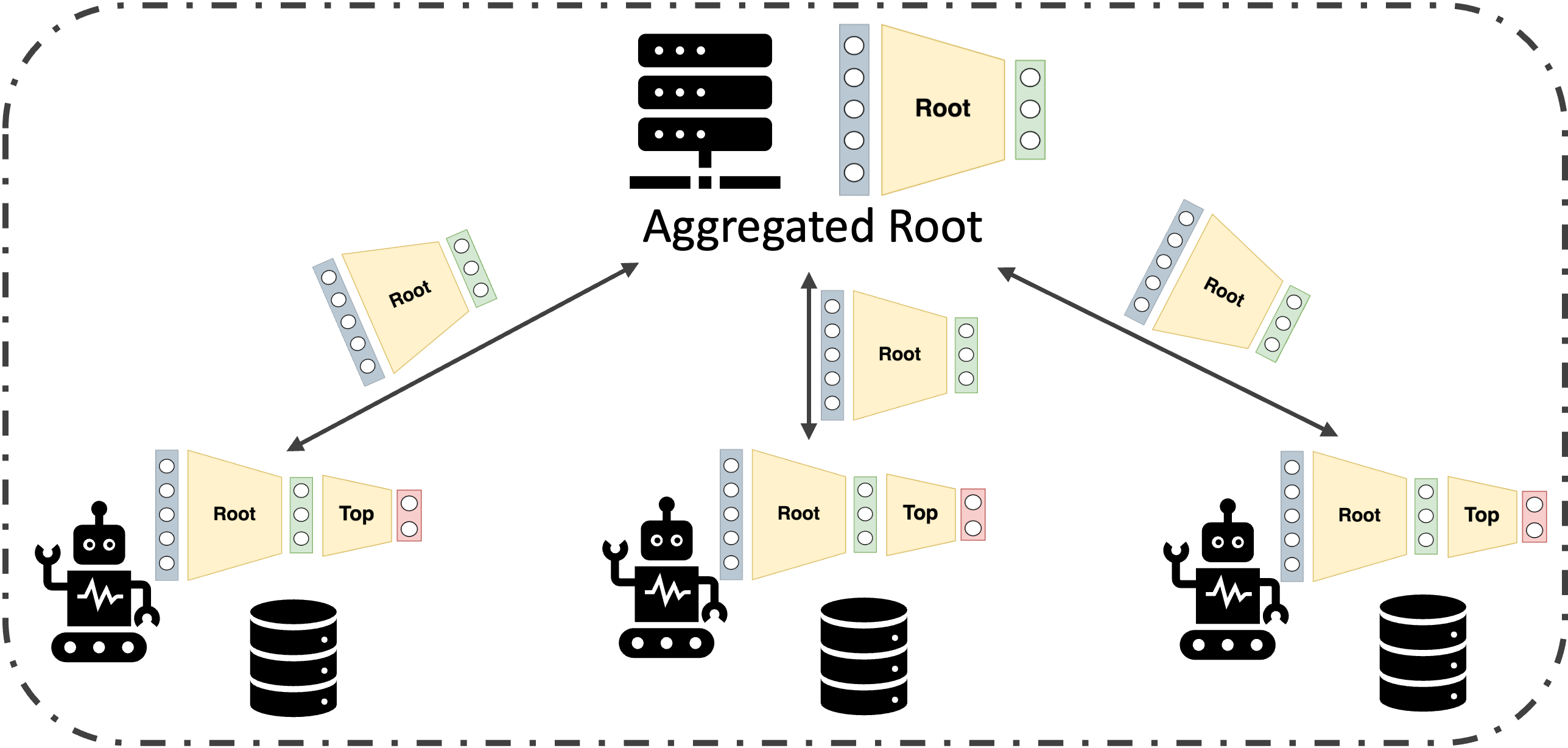}
    \caption{\textbf{\acf{FedRoot}}: Local model split into (i)~\textit{Root} for feature extraction and (ii) \textit{Top} for task-based learning. Only model \textit{Root} is aggregated across clients while \textit{Top} remains local.}
    \label{fig:fedroot}
\end{figure}

\subsection{FedRoot: Federated Root for Feature Aggregation}



To ensure an efficient learning process by avoiding data leakage, we propose \acf{FedRoot} averaging (see Fig.~\ref{fig:fedroot}) as a novel model aggregation \ac{FL} strategy that splits client learning models into two modules:

\subsubsection{Root (R)} The \textit{Root} constitutes the feature extraction layers of the learning model. In our implementation, the convolutional layers constitute the \textit{root} of the model to learn meaningful and relevant features from input images. This is to ensure the \textit{root} learns efficient and descriptive scene embeddings that can help summarise each scene:
\begin{equation}
R = f_{\theta_R}(x),
\end{equation}
\noindent where $f_{\theta_R}$ represents the feature extraction layers (for example, convolutional layers) with parameters $\theta_R$ and the input data  $x$ (for example, scene images).
\subsubsection{Top (T)} The \textit{Top} of the model constitutes the task-based learning (classification or regression) \acf{FC} layers. Using the \textit{Root}-extracted features, the \textit{top} of the model is used for task-based predictions to predict the social appropriateness of different robot actions: 
\begin{equation}
T = h_{\theta_T}(R),
\end{equation}
\noindent where~$h_{\theta_T}$~represents the task-based layers with parameters~$\theta_T$.

Splitting model learning into two parts allows robots to effectively share their learning with other clients while also protecting end-user privacy. 
The entire model $h_{\theta_T}(f_{\theta_R}(X))$ is trained in an end-to-end manner; still, only the \textit{root} from individual client models gets aggregated across clients over repeated aggregation rounds: $Agg(\theta_{R,1}, \theta_{R,2}, \dots, \theta_{R,N})$, where $N$ is the number of clients. This allows the clients to improve their feature embeddings, making them more robust and benefiting from the diversity of data settings experienced by each individual client. Yet, client-specific data, as well as task-based learning $\theta_T$ from the data, that is, end-user preferences on social appropriateness of individual robot behaviours, is kept `\textit{strictly local}'. Since \ac{FedRoot} is proposed as a weight aggregation strategy, we adapt different \ac{FL} approaches explained in Section~\ref{sec:fl} to utilize \ac{FedRoot} and use these methods as our benchmark models. 

\subsection{FedLGR: Federated Latent Generative Replay}

\begin{figure}[t!]
    \centering
    \includegraphics[width=\columnwidth]{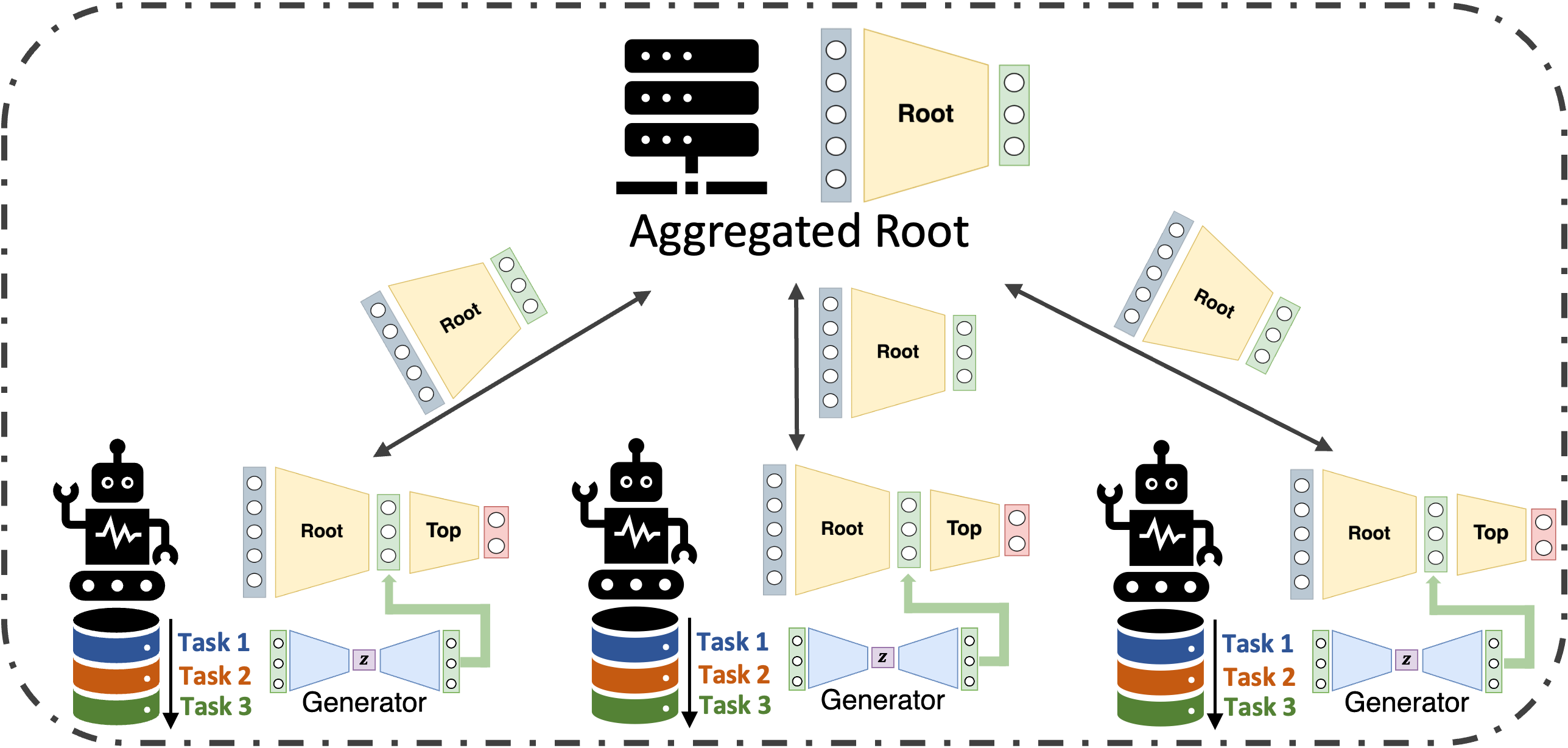}
    \caption{\textbf{\acf{FedLGR}}: Local model split into (i)~\textit{Root} for feature extraction, (ii)~\textit{Top} for task-based learning and (iii)~\textit{Generator} for \textit{pseudo-rehearsal}. Only \textit{Root} is aggregated across clients while \textit{Top} and \textit{Generator} remain local.}
    \label{fig:fedlgr}
\end{figure}

Under \ac{FCL} learning settings, clients not only need to aggregate the learning in a centralised global model but also need to incrementally learn on the acquired `local data', ensuring they do not forget past knowledge. To achieve this, we propose \acf{FedLGR} (see Fig.~\ref{fig:fedlgr}) as a novel \ac{FCL} strategy that applies \acs{LGR}-based pseudo-rehearsal~\cite{Stoychev2023LGR} under \textit{federated learning} settings making use of \ac{FedRoot}-based weight aggregation. 

\acs{FedLGR} uses a \textit{scholar}-based architecture~\cite{Stoychev2023LGR} that consists of three modules: (i)~a \textit{Root} to extract latent feature representations, (ii)~a \textit{Top} to learn task-discriminative information, and (iii)~a \textit{Generator} (Variational Autoencoder) model to reconstruct and rehearse \textit{latent features}. Under \ac{FedLGR}, each client locally uses \acs{LGR} to \textit{continually} and \textit{incrementally} to learn the social appropriateness of robot actions under different contextual settings. This learning is then shared with other clients by aggregating \textit{Root} weights using \ac{FedRoot}. At each time step, the learning for each client \textit{locally} consists of the following steps:

\begin{algorithm}[t!]
\caption{Federated Latent Generative Replay algorithm.}
\label{alg:fedlgr}
\begin{algorithmic}[1]
\For{aggregation round $1 \dots r$}
    \For{input $x$ and target $y$ from current task}
        \State Extract root latent features, $R = f_{\theta_r}(x)$ 
        \State Obtain pseudo-samples $\langle R',T'\rangle$ for old tasks
        \State Update generator $g$ with $(R \cup R^\prime)$ 
        \State Update top $h_{\theta_T}$ using $\langle R, y\rangle$ and $\langle R', T'\rangle$ 
        \State Update root $f_{\theta_R}$ using current task data $x$
    \EndFor
    \State Aggregate root weights across clients:
    \[
    \theta_R^{\text{global}} = Agg(\theta_{R,1}, \theta_{R,2}, \dots, \theta_{R,N})
    \]
    \State Distribute aggregated root weights to clients
\EndFor
\end{algorithmic}
\label{overall_algo}
\end{algorithm}
\begin{enumerate}[\hspace{-0.3cm}(1)]

    \item [a)] The \textit{root} and \textit{top} for the \textit{scholar} are updated with the current task's data, however at different learning rates, $\eta_R$ and $\eta_T$ respectively such that $\eta_R << \eta_T$. This ensures new data does not completely \textit{overwrite} previously seen information. This is important as \ac{LGR}~\cite{Stoychev2023LGR} is based on the assumption that if the latent-space distribution of \textit{root}-extracted features remains relatively static between model updates, the extracted features can be effectively used as a proxy of the input data to rehearse past knowledge~\cite{Pellegrini2020Latent}.
    
    \item [b)] For training the \textit{generator}, the \textit{Root}-extracted feature representations $R$ for the current task's data are interleaved with \textit{generated} features $R^\prime$ for all previously seen tasks:
    \begin{equation}
        G = g (R \cup R^\prime).
    \end{equation}Training the \textit{generator} on both $R$ and $R^\prime$ (only $R$ is used for Task~$1$) ensures that the updated \textit{generator} encodes both new and old tasks. $R^\prime$ is passed to the \textit{top} to obtain labels $T^\prime = h_{\theta_T}(R^\prime)$.
    
    \item [c)] Once the \textit{generator} is updated, the \textit{top} of the \textit{solver} is updated using the current task data $\langle R, y \rangle$, interleaved with `labelled' \textit{latent pseudo-samples} $\langle R^\prime, T^\prime \rangle$ generated for previously seen tasks. The \textit{root} is then updated with $\eta_R$ using only the current task's data. 
\end{enumerate}

\noindent Training on each task consists of several rounds of weight aggregations ($r=5$ per task), where \ac{FedRoot} is used to aggregate \textit{root} module weights across clients. The \textit{top} and the \textit{generator} for each client are kept `strictly local', ensuring task-discriminative information is never shared. The overall methodology is provided in Algorithm \ref{overall_algo}.



\section{Experiments}



\subsection{MANNERS-DB Dataset}
For our proof-of-concept evaluations on \ac{FL} and \ac{FCL}, we explore the  MANNERS-DB dataset~\cite{tjomsland2022mind} that consists of Unity-generated scenes of the Pepper robot co-inhabiting a living room (see Figure~\ref{fig:mannerdb}) with other humans and animals under different social settings. MANNERS-DB has been explored for \ac{CL}-based~\cite{tjomsland2022mind} and zero-shot~\cite{Zhang2023LLM} learning, however, not for \ac{FCL}. For each scene, the robot can perform $8$ different actions, that is, \textit{vacuuming, mopping, carrying warm food, carrying cold food, carrying big objects, carrying small objects, carrying drinks and cleaning/starting conversations} either \textit{within a circle of influence} or in the \textit{direction of operation} (see Figure~\ref{fig:mannerdb}). Crowd-sourced annotations are provided for the social appropriateness of each of these actions for every scene ($\approx 1000$ scenes), labelled on a $5$-point Likert scale, ranging from \textit{very inappropriate} to \textit{very appropriate}. 
For both \ac{FL} and \ac{FCL} evaluations, the data was split into training and test splits in the ratio of $3:1$. The training data was further split amongst the different clients (either $2$ or $10$) with a shared test set used for evaluation. For \ac{FCL} evaluations, instead of incrementally learning separate actions, the training set of individual clients was split into two tasks (each containing a balanced number of samples), that is, samples depicting the robot operating with a circle (\textbf{Task~$\bm{1}$}) and in the direction of the arrow (\textbf{Task~$\bm{2}$}). It is important to highlight that the \ac{FedLGR} methodology is task-agnostic and adaptable to diverse incremental learning tasks without requiring any architecture modifications.

For training the models, \textit{normalised} 
RGB images were used, resized to $(128 \times 128 \times 3)$ due to computational restrictions of training multiple clients in parallel on GPU. Since MANNERS-DB is a relatively small dataset consisting of $\approx 1000$ images, we evaluate the different approaches \textit{with} and \textit{without} data augmentation. Although various methods could be used for data augmentation, we adopted a straightforward approach to evaluate whether data augmentation can offer meaningful contributions.  Specifically, we used random ($p=0.5$) horizontal flipping and random ($p=0.5$) rotation (up to $10^\circ$) in either direction to augment the dataset. For a fair comparison, results were presented individually for \textit{without} and \textit{with} augmentation comparisons.

\subsection{Implementation Details}
For each approach, a \acf{CNN}-based model was used consisting of two parts: (i)~a \textit{Conv} module that uses the popularly used MobileNet-V2~\cite{Sandler2018MobileNet} backbone to extract scene features, followed by an \textit{AdaptiveAvgPool} layer for dimensionality reduction, and (ii)~an \textit{FC} module where the $1280-$d \textit{flattened} scene features are passed through a \ac{FC} layer consisting of $32$ units followed by the $8-$unit output layer (each unit for per robot action). Each layer in the \textit{FC} module was followed by \textit{BatchNorm} layer and uses a \textit{linear} activation with the output layer predicting the \textit{social appropriateness} of all $8$ robot actions. All models were trained using the \textit{Adam} optimiser ($\beta_1$$=$$0.9$, $\beta_2$$=$$0.999$) with \textit{learning rate} of $lr$$=$$1e^-3$. All methods were implemented using the \textit{Flower}
and PyTorch Python libraries. The \ac{FL} and \ac{FCL} experiments were run for $2-10$ clients, attuned to a potential real-world evaluation to be conducted in the future using physical robots. For brevity, results for only $2$ and $10$ clients are presented. 

Model hyper-parameters and regularisation coefficients for \ac{FCL} approaches were optimised using a grid search. For without augmentation experiments, the \textit{BatchSize} was empirically set to $16$ for \textit{two clients} and $8$ for \textit{ten clients} due to the overall low number of samples per client. When using with augmentation, \textit{BatchSize} was set to $16$. All models were trained using an \ac{MSE} loss, computed as an average across the $8$ actions. Each client agent trained on its own split of the training data, but a shared test set was used for model evaluation. We reported the average \ac{MSE} loss averaged across the $8$ actions and all clients.


        

\begin{table}[t!]
    \caption{Federated Learning results on the test set of MANNERS-DB Dataset for two (left) and ten (right) clients. \textbf{Bold} values denote best while [bracketed] denote second-best values.}
    \label{tab:FL}
    \setlength{\tabcolsep}{2.4pt}

    \centering
    \renewcommand{\arraystretch}{0.93}

    \begin{tabular}{l l|ccc|ccc}\toprule
        & & \multicolumn{3}{c|}{\textit{Two Clients}} &   \multicolumn{3}{c}{\textit{Ten Clients}}\\\midrule

        &Method          & Loss $\blacktriangledown$             & RMSE $\blacktriangledown$             & PCC   $\blacktriangle$            & Loss $\blacktriangledown$             & RMSE  $\blacktriangledown$            & PCC $\blacktriangle$              \\\midrule

        & & \multicolumn{6}{c}{\textit{W/O Augmentation}}\\\midrule
        & FedAvg~\cite{Li2020FedAvg}  & 0.264 & 0.510 & [0.508] & 0.220 & 0.467 & 0.487 \\
        & FedBN~\cite{Li2021FedBN} & 0.310 & 0.553 & 0.486 & 0.226 & 0.471 & 0.486 \\
        & FedProx~\cite{Li2020FedProx} & [0.262] & [0.507] & \textbf{0.513} & [0.211] & [0.459] & \textbf{0.517} \\
        & FedOpt~\cite{Reddi2021FedOpt} & \textbf{0.245} & \textbf{0.492} & 0.501 & \textbf{0.210} & \textbf{0.458} & [0.492] \\
        & FedDistill~\cite{Jiang2020FedDistill} & 0.276 & 0.522 & 0.475 & 0.233 & 0.481 & 0.470 \\ \midrule
        \multirow{5}{*}{\rotatebox{90}{Ours}} & FedRootAvg & 0.276 & 0.522 & 0.482 & 0.231 & 0.477 & 0.459 \\
        & FedRootBN & 0.272 & 0.518 & 0.498 & 0.263 & 0.510 & 0.420 \\
        & FedRootProx & 0.264 & 0.511 & 0.481 & 0.240 & 0.486 & 0.447 \\
        &FedRootOpt & 0.285 & 0.530 & 0.473 & 0.240 & 0.486 & 0.460 \\
        &FedRootDistill & 0.294 & 0.539 & 0.489 & 0.305 & 0.550 & 0.363 \\\midrule
        
        & & \multicolumn{6}{c}{\textit{W/ Augmentation}}\\\midrule
        
        &FedAvg~\cite{Li2020FedAvg} & 0.192 & 0.435 & 0.547 & 0.178 & [0.420] & \textbf{0.586} \\
        & FedBN~\cite{Li2021FedBN} & 0.205 & 0.447 & [0.554] & 0.188 & 0.429 & 0.567 \\
        &FedProx~\cite{Li2020FedProx} & 0.208 & 0.451 & 0.536 & \textbf{0.173} & \textbf{0.412} & [0.581] \\
        &FedOpt~\cite{Reddi2021FedOpt} & 0.204 & 0.450 & 0.506 & [0.177] & [0.420] & 0.568 \\
        &FedDistill~\cite{Jiang2020FedDistill} & 0.221 & 0.468 & 0.538 & 0.205 & 0.449 & 0.560 \\ \midrule
        \multirow{5}{*}{\rotatebox{90}{Ours}}&FedRootAvg& 0.192 & 0.433 & 0.547 & 0.198 & 0.440 & 0.563 \\
        &FedRootBN & 0.197 & 0.442 & \textbf{0.558} & 0.194 & 0.439 & 0.539 \\
        &FedRootProx & \textbf{0.186} & \textbf{0.427} & 0.541 & 0.196 & 0.439 & 0.552 \\
        &FedRootOpt & [0.190] & [0.432] & 0.528 & 0.206 & 0.450 & 0.549 \\
        &FedRootDistill & 0.225 & 0.469 & 0.525 & 0.222 & 0.465 & 0.539 \\\bottomrule
    \end{tabular}
\end{table}

\begin{table*}[ht!]
\centering
  \caption{Average CPU usage (s) and GPU usage (\%) per client, per aggregation round for Federated Learning experiments.}
  \label{tab:flusage}
  {
\setlength{\tabcolsep}{2.0pt}
\renewcommand{\arraystretch}{0.93}

\begin{tabular}{r|cc|cc|cc|cc|cc} \toprule
    Metric & FedAvg & FedRootAvg & FedBN & FedRootBN & FedProx & FedRootProx & FedOpt & FedRootOpt & FedDistill & FedRootDistill\\
    \midrule
    CPU Usage (s) $\blacktriangledown$ & 19.06 & \textbf{2.55} (\textcolor{green}{$\blacktriangledown$} 86.6\%)  & 18.4 & \textbf{2.50} (\textcolor{green}{$\blacktriangledown$} 86.4\%) & 18.98 & \textbf{2.54} (\textcolor{green}{$\blacktriangledown$} 86.6\%) & 18.61 & \textbf{2.46} (\textcolor{green}{$\blacktriangledown$} 86.7\%) & 19.46 &  \textbf{2.61} (\textcolor{green}{$\blacktriangledown$} 86.4\%) \\\midrule

    GPU Usage (\%) $\blacktriangledown$ & 0.63 & \textbf{0.24} (\textcolor{green}{$\blacktriangledown$} 61.9\%)  & 0.93 & \textbf{0.27} (\textcolor{green}{$\blacktriangledown$} 70.9\%) & 0.25 & \textbf{0.15} (\textcolor{green}{$\blacktriangledown$} 40.0\%) & 0.69 & \textbf{0.19} (\textcolor{green}{$\blacktriangledown$} 72.4\%) & 0.10 & \textbf{0.07} (\textcolor{green}{$\blacktriangledown$} 30.0\%) \\     \bottomrule
    

    
\end{tabular}
}
\end{table*}

\begin{table*}[ht!]
\centering
    \vspace{4mm}
  \caption{Average CPU usage (s) and GPU usage (\%) per client, per aggregation round for Federated Continual Learning experiments.}
  \label{tab:fclusage}
  {
\setlength{\tabcolsep}{0.8pt}
\renewcommand{\arraystretch}{0.94}

  \begin{tabular}{r|cc|cc|cc|cc|cc|c} \toprule
    Metric & \makecell[c]{FedAvg\\EWC} & \makecell[c]{FedRoot\\EWC} & \makecell[c]{FedAvg\\EWCOnline} & \makecell[c]{FedRoot\\EWCOnline} & \makecell[c]{FedAvg\\MAS} & \makecell[c]{FedRoot\\MAS} & \makecell[c]{FedAvg\\SI} & \makecell[c]{FedRoot\\SI} & \makecell[c]{FedAvg\\NR} & \makecell[c]{FedRoot\\NR} & FedLGR \\
    \midrule
    CPU Usage (s) $\blacktriangledown$ & 24.17 & \textbf{3.87} (\textcolor{green}{$\blacktriangledown$} 83.9\%)  & 25.05 & \textbf{4.97} (\textcolor{green}{$\blacktriangledown$} 80.2\%) & 22.41 & \textbf{5.36} (\textcolor{green}{$\blacktriangledown$} 76.1\%)  & 20.32 & \textbf{3.91} (\textcolor{green}{$\blacktriangledown$} 80.7\%) & 25.97 & \textbf{4.05} (\textcolor{green}{$\blacktriangledown$} 84.4\%) & \textbf{4.06}
    \\ \midrule

    GPU Usage (\%) $\blacktriangledown$ & 0.57 & \textbf{0.04} (\textcolor{green}{$\blacktriangledown$} 92.9\%) & 0.13 & \textbf{0.03} (\textcolor{green}{$\blacktriangledown$} 76.9\%) & 0.04 & \textbf{0.01} (\textcolor{green}{$\blacktriangledown$} 75.0\%) & 0.10 & \textbf{0.05} (\textcolor{green}{$\blacktriangledown$} 50.0\%) & 0.06  & \textbf{0.04} (\textcolor{green}{$\blacktriangledown$} 33.3\%)  & \textbf{0.02} 
    \\ \bottomrule
    
    
    
    

        
\end{tabular}

}
\end{table*}

\subsection{Evaluation Metrics}
Predicting the social-appropriateness for each robot action requires transforming \ac{FL} and \ac{FCL} methods to use regression-based objectives, using the following metrics:

\begin{itemize}

    \item \textit{\acf{RMSE}:} We report the \ac{RMSE} values, which are calculated on the test set, averaged across the $8$ actions across all clients. \ac{RMSE} scores are used to compare our results with the baseline provided with the MANNERS-DB dataset~\cite{tjomsland2022mind}.

    \item \textit{\acf{PCC}:} In addition to an absolute metric such as the \ac{RMSE}, we also evaluate the predicted social appropriateness values relative to the ground truth using the \ac{PCC}~\cite{benesty2009pearson} scores. \ac{PCC} scores are calculated on the test set individually for each of the $8$ actions for each client. Average scores across all actions and all clients are reported. 

    \item \textit{CPU Usage (s):} To evaluate the resource efficiency of \ac{FL} and \ac{FCL} approaches, CPU usage is reported as the CPU time (in seconds) allocated to training each client on average. This is calculated as the CPU time utilised by each client across all rounds of weight aggregation. 
    
    \item \textit{GPU Usage (\%):} Similar to CPU usage, GPU usage is also important to evaluate the resource efficiency of the proposed methods. Since we have a single GPU simulation set-up, we report the percentage ($\%$) of the GPU allocated per client on average. GPU usage is also calculated using Nvidia's \verb|nvidia-smi| tool, logged at different intervals during the training process, and \textit{averaged} over the entire training time for each client.

\end{itemize}

\section{Results}

\subsection{Federated Learning Benchmark}
Table~\ref{tab:FL} presents the \ac{FL} benchmark results on the MANNERS-DB dataset comparing popular \ac{FL} strategies and their \ac{FedRoot}-based adaptations for \textit{two} and \textit{ten} clients, both without and with data augmentation, respectively. 
For without data augmentation experiments with \textit{two} and \textit{ten clients}, FedOpt and FedProx emerge as the best-performing approaches, on average. The `general optimisation framework' of FedOpt, with \textit{Adam}-based client and server optimisers, is able to efficiently aggregate learning across clients, resulting in the robust performance of the model. FedProx, on the other hand, adds a proximal term to \ac{FedAvg} and updates the objective for each client.
For the proposed \acs{FedRoot} variants of the compared \ac{FL} approaches, as task-discriminative \textit{top} weights are kept \textit{strictly local}, we see a slight decrease in model performance. This is primarily due to the small size of the MANNERS-DB dataset, which does not have enough data samples for the FedRoot to optimise model performance, given that only feature extracting \textit{root} is aggregated. However, the critical contribution of \acs{FedRoot} lies in offering a \textit{significant reduction} in CPU and GPU usage per client, per aggregation round, compared to the original methods, as can be seen in Table~\ref{tab:flusage}. 

As Table~\ref{tab:FL} demonstrates, data augmentation has a net positive impact on all models, with all approaches reporting better metrics across evaluations. With the larger amount of data available per client, we see that \acs{FedRoot}-based methods improve upon their counterparts across all metrics for the experiments with \textit{two clients}. Similar to without augmentation experiments, using \textit{Adam}-based client and server optimisation as well as adding a proximal term to the model learning objective 
enhance performances, yielding the best ones 
for \acs{FedRoot}Opt and \acs{FedRoot}Prox. 
On the other hand, data available per client still remains relatively low when split across \textit{ten clients}, even when using data augmentation. Still, \acs{FedRoot}-based approaches demonstrate an overall improvement compared to results without augmentation and excel in resource efficiency while maintaining competitive performance for \textit{ten clients}. 



\begin{table*}
    \caption{Federated Continual Learning results on the MANNERS-DB Dataset for two (left) and ten (right) clients. `After Task 1' shows the results on the test set of Task 1, and `After Task 2' shows the results on the test set of both tasks. \textbf{Bold} values denote best while [bracketed] denote second-best values.}
    \label{tab:FCL}
    \centering
    \setlength{\tabcolsep}{3.0pt}
    \renewcommand{\arraystretch}{0.93}

    \begin{tabular}{l lccc|ccc|ccc|ccc} \toprule
    & &\multicolumn{6}{c|}{\textit{Two Clients}} &   \multicolumn{6}{c}{\textit{Ten Clients}}\\\midrule

    & &\multicolumn{3}{c|}{After Task 1}&\multicolumn{3}{c|}{After Task 2} &\multicolumn{3}{c|}{After Task 1}&\multicolumn{3}{c}{After Task 2} \\ 
    
    & Method & Loss $\blacktriangledown$             & RMSE $\blacktriangledown$             & PCC   $\blacktriangle$  & Loss $\blacktriangledown$             & RMSE $\blacktriangledown$             & PCC   $\blacktriangle$   & Loss $\blacktriangledown$             & RMSE $\blacktriangledown$             & PCC   $\blacktriangle$   & Loss $\blacktriangledown$             & RMSE $\blacktriangledown$             & PCC   $\blacktriangle$   \\\midrule
    & \multicolumn{12}{c}{\textit{W/O Data-augmentation}}\\ \midrule
     & FedAvg - EWC~\cite{kirkpatrick2017overcomingEWC}& 0.705 & 0.826 & 0.488 & 0.911 & 0.944 & 0.347 & 0.364 & 0.591 & 0.325 & 0.393 & 0.617 & 0.278\\
    
    & FedAvg - EWCOnline~\cite{schwarz2018progressEWCOnline} 
    & 0.488 & 0.685 & 0.469 & 0.381 & 0.608 & 0.384 
    & [0.328] & [0.566] & 0.373 & [0.359] & [0.589] & 0.321\\
    
    &FedAvg - MAS~\cite{aljundi2018memoryMAS}
    & 0.568 & 0.740 & 0.570 & 0.363 & 0.597 & 0.357 
    & 0.421 & 0.641 & [0.494] & 0.464 & 0.674 & 0.357\\
    
    & FedAvg - SI~\cite{zenke2017continualSI} & 0.488 & 0.686 & [0.571] & 0.675 & 0.811 & 0.343  & 0.451 & 0.665 & 0.323 & 0.376 & 0.598 & 0.306\\
    
    & FedAvg - NR~\cite{Hsu18_EvalCLNR} 
    & 0.403 & 0.634 & 0.491 & [0.296] & [0.538] & 0.333 
    & 0.376 & 0.600 & 0.447 & 0.408 & 0.628 & 0.341\\\midrule
    
    \multirow{5}{*}{\rotatebox{90}{Ours}} & FedRoot - EWC 
    & 0.433 & 0.652 & 0.472 & 0.998 & 0.963 & 0.382 
    & 9.320 & 3.050 & 0.336 & 8.950 & 2.990 & 0.335\\
    
    &FedRoot - EWCOnline 
    & 0.614 & 0.778 & 0.436 & 0.447 & 0.659 & 0.400 
    & 9.170 & 3.030 & 0.330 & 9.560 & 3.090 & [0.376]\\
    
    &FedRoot - MAS
    & [0.397] & [0.627] & 0.429 & 0.569 & 0.740 & 0.422
    & 8.860 & 2.980 & 0.363 & 8.750 & 2.960 & 0.369\\
    
    &FedRoot - SI
    & 0.968 & 0.984 & 0.441 & 0.840 & 0.893 & 0.418 
    & 9.360 & 3.060 & 0.417 & 9.120 & 3.020 & 0.369\\
    
    &FedRoot - NR 
    & 0.762 & 0.870 & 0.325 & \textbf{0.287} & \textbf{0.533} & [0.444] 
    & 0.446 & 0.656 & 0.462 & 0.482 & 0.677 & 0.339\\ \midrule

    &FedLGR (ours) 
    & \textbf{0.252} & \textbf{0.499} & \textbf{0.599} & 0.465 & 0.677 & \textbf{0.534} 
    & \textbf{0.299} & \textbf{0.539} & \textbf{0.515} & \textbf{0.320} & \textbf{0.553} & \textbf{0.421}\\\midrule

    & & \multicolumn{12}{c}{\textit{W/ Data-augmentation}}\\ \midrule
    &FedAvg - EWC~\cite{kirkpatrick2017overcomingEWC} & 0.669 & 0.807 & 0.420 & 0.348 & 0.584 & 0.395 & 0.397 & 0.624 & 0.472 & 0.433 & 0.653 & 0.456\\
    
    & FedAvg - EWCOnline~\cite{schwarz2018progressEWCOnline} & 0.495 & 0.702 & 0.438 & 0.629 & 0.788 & 0.376 & [0.319] & [0.561] & 0.365 & [0.329] & [0.579] & 0.363\\
    
    & FedAvg - MAS~\cite{aljundi2018memoryMAS} & 0.267 & 0.514 & \textbf{0.601} & 0.665 & 0.814 & 0.394 & 0.343 & 0.580 & 0.480 & 0.354 & 0.587 & 0.328 \\
    
    & FedAvg - SI~\cite{zenke2017continualSI} & 0.280 & 0.526 & [0.546] & [0.328] & [0.568] & 0.444& 0.471 & 0.682 & 0.418 & 0.435 & 0.654 & 0.426\\
    
    & FedAvg - NR~\cite{Hsu18_EvalCLNR} & 0.450 & 0.669 & 0.531 & 0.710 & 0.839 & 0.451 & 0.392 & 0.618 & 0.425 & 0.490 & 0.698 & 0.368\\ \midrule
    
    \multirow{5}{*}{\rotatebox{90}{Ours}}& FedRoot - EWC 
    & 0.736 & 0.836 & 0.529 & \textbf{0.327} & \textbf{0.569} & 0.463 & 0.441 & 0.648 & [0.543] & 0.449 & 0.655 & [0.467] \\
    
    & FedRoot - EWCOnline 
    & 0.517 & 0.712 & 0.398 & 0.446 & 0.661 & 0.363 & 0.688 & 0.805 & 0.446 & 0.612 & 0.761 & 0.405\\
    
    & FedRoot - MAS & [0.265] & [0.511] & 0.492 & 0.616 & 0.782 & 0.436 & 0.797 & 0.851 & 0.542 & 1.030 & 0.930 & 0.439\\
    
    & FedRoot - SI & 0.388 & 0.596 & 0.405 & 0.735 & 0.821 & [0.478] & 0.481 & 0.688 & 0.323 & 0.521 & 0.708 & 0.439\\
    
    & FedRoot - NR 
    & 0.781 & 0.875 & 0.491 & 0.368 & 0.604 & 0.354 
    & 0.539 & 0.728 & 0.268 & 0.409 & 0.635 & 0.447\\ \midrule
    
    & \ac{FedLGR} (ours) 
    & \textbf{0.230} & \textbf{0.478} & 0.531 & 0.493 & 0.699 & \textbf{0.479} 
    & \textbf{0.288} & \textbf{0.528} & \textbf{0.563} & \textbf{0.317} & \textbf{0.560} & \textbf{ 0.482}\\\bottomrule
\end{tabular}

\end{table*}

\subsection{Federated Continual Learning Benchmark}
Table~\ref{tab:FCL} presents the \ac{FCL} results on the MANNERS-DB dataset adapting \ac{FedAvg} and \acs{FedRoot}-based weight aggregation strategies to use \ac{CL}-based learning objectives to mitigate forgetting under incremental settings. Evaluation metrics are reported after training and testing on the data-split depicting the robot operating \textit{within the circle of influence} \textbf{(Task~1)} and training on the data-split depicting the robot \textit{in the direction of operation} \textbf{(Task~2)} and testing on both the splits.  Similar to the \ac{FL} benchmark, we compare the different \acs{FCL} strategies for \textit{two} and \textit{ten} clients as well as without and with data augmentation. As \textit{federated averaging} of weights struggles with learning under non-\acs{i.i.d} data settings~\cite{Li2021FedBN}, extending these methods with \ac{CL}-based objectives allows us to evaluate how they can objectively contribute towards maintaining model performance when learning \textit{incrementally}. 

Table~\ref{tab:FCL} demonstrates that the proposed \acs{FedLGR} approach performs the best across all evaluations, especially for scarce data, for example, while learning across \textit{ten} clients. \ac{FedLGR} is able to efficiently use pseudo-rehearsal of features to maintain model performance after learning Task~2.
Additionally, \ac{FedRoot}-based approaches are able to outperform their \ac{FedAvg}-based counterparts across most evaluations. However, similar to \ac{FL} evaluations, \acs{FedRoot}-based methods are affected by the scarcity of data, for instance, the evaluations across \textit{ten} clients without data augmentation, resulting in significantly worse loss and \acs{RMSE} scores. Data augmentation has a net positive impact on model performances, especially for \ac{FedRoot}-based methods. Importantly, using \ac{FedRoot}-based weight aggregation results in a sizeable (up to \textit{84\%} for CPU usage and \textit{92\%} for GPU usage) reduction in computational expense of running these methods, per client, per aggregation round (see Table~\ref{tab:fclusage}). This can be particularly beneficial for resource-constrained applications such as service robots helping people in the real world. 

\section{Discussion and Conclusion}
Social robots operating in dynamic real-world settings can benefit from federated learning mechanisms where, learning from and adapting towards their unique environmental and data conditions, they can also share their learning with other robots, benefiting from each others' experiences. \ac{FL}-based approaches enable such a learning paradigm for robots to aggregate model updates across individual agents into a global model while ensuring that end-user privacy is preserved. However, the resource constraints inhibit such learning settings, resulting in most existing methods opting for \textit{centralised} learning, where end-user devices only collect data for a centralised model to be trained, in isolation, for later application \textit{in-the-wild}. The proposed \acf{FedRoot} weight aggregation strategy addresses these challenges by splitting each client's learning model into \textit{aggregatable} feature extraction layers, that is, model \textit{root} and \textit{private} task-relevant \textit{top} layers that learn to predict the social appropriateness of different robot actions. The \ac{FL} benchmark results (see Table~\ref{tab:FL} and Table~\ref{tab:flusage}) highlight the competitive performance of \acs{FedRoot}-based approaches with sizeable reductions in CPU and GPU usage for each client. These findings are specifically important for human-robot collaboration and service robots, where resource efficiency is critical for successfully completing the task, for example, when robots carry big objects or clean large spaces.

Furthermore, real-world applications require robots to learn incrementally,  
for instance, in the form of developing novel capabilities or applying existing capabilities under novel contextual settings. In this work, we explore the latter, where robots need to learn the social appropriateness of different actions depending upon the context in which they are operating, that is, operating \textit{within the circle of influence} or \textit{in a particular direction of operation}. We adapt the proposed \acs{FedRoot}Avg weight aggregation strategy and extend it by adapting popular \ac{CL}-based learning objectives presenting a novel \ac{FCL} benchmark on the MANNERS-DB dataset, as shown in Table~\ref{tab:FCL} and Table~\ref{tab:fclusage}). In particular, the proposed \acf{FedLGR} approach is seen to outperform other methods across all evaluations. It implements a local \textit{generator} for efficient pseudo rehearsal of latent features for mitigating forgetting, where the \textit{top} learns task-relevant information. 
Our work contributes significantly to the fields of Federated Learning and robotics, offering promising avenues for the development of socially intelligent machines capable of learning and adapting in a decentralized and resource-efficient manner.

Our work presents novel \ac{FL} and \ac{FCL} benchmarks for learning the social appropriateness of high-level robot behaviours in simulated home settings, and future work can investigate the usability of our system in the real world with a user study where multiple robots operate in a household and share their knowledge with each other. 
However, MANNERS-DB currently has the limitation of focusing solely on living room scenes. Future research could extend this dataset to include a wider variety of environments (for example, kitchen, bedroom, office), different robot embodiments (for example, Pepper, Nao, PR2), and diverse points-of-view (robot-centric and scene-centric). Such expansions would enable a more comprehensive evaluation of \acs{FL} and \acs{FCL} across multiple clients, offering a more generalizable assessment of large-scale \textit{federated} applications for robots in diverse real-world scenarios. Future work could also investigate the scalability of \ac{FedRoot} and \ac{FedLGR} in broader tasks beyond social appropriateness.

\begin{acronym}
    \acro{AI}{Artificial Intelligence}

    \acro{CF}{Catastrophic Forgetting}
    \acro{CL}{Continual Learning}
    \acro{Class-IL}{Class-Incremental Learning}
    \acro{CNN}{Convolutional Neural Network}

    \acro{DGR}{Deep Generative Replay}
    \acro{DNN}{Deep Neural Network}
    
    \acro{EWC}{Elastic Weight Consolidation}

    \acro{FL}{Federated Learning}
    \acro{FC}{Fully Connected}
    \acro{FCL}{Federated Continual Learning}
    \acro{FedAvg}{Federated Averaging}
    \acro{FedRoot}{Federated Root}
    \acro{FedLGR}{Federated Latent Generative Replay}

    \acro{HRI}{Human-Robot Interaction}

    \acro{i.i.d}{\textit{independent and identically distributed}}

    \acro{LGR}{Latent Generative Replay}
    \acro{lr}{learning rate}

    \acro{MAS}{Memory Aware Synapses}
    \acro{ML}{Machine Learning}
    \acro{MLP}{Multilayer Perceptron}
    \acro{MSE}{Mean Squared Error}
    
    \acro{NC}{New Concepts}
    \acro{NI}{New Instances}
    \acro{NIC}{New Instances and Concepts}
    \acro{NR}{Naive Rehearsal}

    \acro{PCC}{Pearson's Correlation Coefficient}
    
    \acro{RL}{Reinforcement Learning}
    \acro{RMSE}{Root Mean Squared Error}

    \acro{SI}{Synaptic Intelligence}
    
    \acro{Task-IL}{Task-Incremental Learning}

\end{acronym}
\bibliographystyle{ieeetr}
\bibliography{main}

\end{document}